\documentclass{article} 
\usepackage{iclr2027_conference,times}

\usepackage{amsmath,amsfonts,bm}

\def\eqref#1{equation~\ref{#1}}

\def\1{\bm{1}}

\DeclareMathAlphabet{\mathsfit}{\encodingdefault}{\sfdefault}{m}{sl}
\SetMathAlphabet{\mathsfit}{bold}{\encodingdefault}{\sfdefault}{bx}{n}

\usepackage{hyperref}
\usepackage{url}
\usepackage{graphicx}
\usepackage{booktabs}
\usepackage{multirow}
\usepackage{pifont}

\newcommand{\cmark}{\ding{51}}
\newcommand{\xmark}{\ding{55}}

\title{TaskIR: Task-Driven Image Restoration via Degradation Adaptation and Task Feedback}

\author{
\parbox{0.98\textwidth}{
\centering
\mbox{Yanjie Tu$^{1}$}\quad
\mbox{Qingsen Yan$^{1,2}$\thanks{Corresponding author: qingsenyan@nwpu.edu.cn}}\quad
\mbox{Axi Niu$^{1}$}\quad
\mbox{Wenxuan Cai$^{1}$}
\\[2pt]
\mbox{Tao Hu$^{1}$}\quad
\mbox{Wei Dong$^{3}$}\quad
\mbox{Haokui Zhang$^{1}$}
\\[4pt]
{\normalfont
$^{1}$Northwestern Polytechnical University \quad
$^{2}$Shenzhen Research Institute of Northwestern Polytechnical University \quad
$^{3}$Xi'an University of Architecture and Technology
}
}
}

\iclrfinalcopy 
\begin{document}

\maketitle

\begin{abstract}
Task-driven image restoration aims to improve both image quality and downstream task performance. However, existing methods predominantly focus on single degradation type and struggle to handle the diverse degradations encountered in real-world scenarios. Different degradations impose distinct restoration demands, and insufficient restoration may leave residual degradations and artifacts that impair object boundaries and semantic cues, thereby compromising downstream task performance. To address these challenges, we propose TaskIR, a two-stage task-driven unified image restoration framework that integrates degradation-adaptive restoration with task feedback refinement. In Stage I, a Degradation Representation Module (DRM) extracts degradation representations, enabling a Degradation-Guided Transformer Block (DGTB) to dynamically modulate feature transformations for adaptive restoration. In Stage II, a Task-to-Restoration Feedback Generation module (TRFG) transforms heterogeneous task features into restoration feedback by modeling task-representation discrepancies associated with the current restoration. Subsequently, a Selective Task Feedback Refinement module (STFR) assesses feedback relevance and selectively refines intermediate restoration features to mitigate interference with well-restored content. Extensive experiments demonstrate that TaskIR achieves competitive restoration quality and downstream task performance across diverse degradations and tasks. The code is available at \href{https://github.com/leoyjTu/TaskIR}{TaskIR}.
\end{abstract}

\section{Introduction}
Image restoration aims to reconstruct high-quality images from degraded observations and provide reliable inputs for subsequent visual analysis. Images captured in real-world scenarios are often affected by various degradations, such as noise~\citep{noise}, blur~\citep{blur}, haze~\citep{haze}, and low illumination~\citep{low}. These degradations not only deteriorate visual quality but may also corrupt task-relevant information, such as object textures, structures, and boundaries, thereby degrading the performance of downstream vision tasks, including image classification~\citep{cls}, semantic segmentation~\citep{seg}, and object detection~\citep{det}. Consequently, task-driven image restoration~\citep{edtr,tasktok,adet-net} has attracted increasing attention, aiming to improve restoration quality while preserving or enhancing information relevant to downstream tasks.
Existing task-driven image restoration methods typically guide the restoration process using task losses, task features, or prediction results and have achieved promising performance in specific scenarios such as super-resolution~\citep{tdsr,sr4ir} and dehazing~\citep{ms-fodn,adet-net}. However, most of these methods are designed for a single degradation, with their network architectures and task-guidance mechanisms often tailored to specific degradation conditions, making them difficult to directly adapt to the diverse and complex degradations encountered in real-world scenarios. In fact, different degradations exhibit distinct corruption patterns and restoration requirements, and their effects on downstream task representations and prediction performance also vary. Therefore, extending task-driven image restoration from single-degradation settings to a unified multi-degradation scenario represents an important direction toward practical deployment in complex real-world environments.

In recent years, all-in-one image restoration~\citep{promptir,survey,moceir,baryir} has attracted increasing attention by handling multiple degradations within a single model and enhancing multi-degradation restoration through mechanisms such as degradation representation, prompt learning, and dynamic experts. However, different degradations have distinct restoration requirements, making it difficult for shared feature transformations to achieve sufficient degradation adaptation, while improved restoration quality does not necessarily lead to better downstream task performance~\citep{task-unirestore}. On the other hand, downstream task features emphasize high-level semantic and discriminative information, which differs from the texture and structural information required for image restoration, and directly using such features for restoration may introduce interference. Moreover, different task feedback signals contribute differently to restoration, and indiscriminately using them to refine restoration features may disrupt already restored content~\citep{tasktok}. Therefore, adapting to the restoration requirements of diverse degradations while effectively utilizing heterogeneous task feedback to improve downstream task performance without compromising restoration quality remains a core challenge for task-driven image restoration.

To address the above challenges, we propose TaskIR, a two-stage task-driven all-in-one image restoration framework. In Stage I, we design a Degradation Representation Module (DRM) to jointly model global degradation characteristics and spatial degradation responses and employ a Degradation-Guided Transformer Block (DGTB) to dynamically modulate feature transformations for degradation-adaptive restoration. In Stage II, we further incorporate downstream task feedback to improve the suitability of restoration results for downstream tasks. Specifically, Task-to-Restoration Feedback Generation (TRFG) adaptively transforms heterogeneous task features and generates targeted task feedback based on discrepancies in the task representations of the current restoration results. Subsequently, Selective Task Feedback Refinement (STFR) assesses the contributions of different feedback signals to the current restoration state and selectively refines restoration features to minimize interference with already restored content. By combining degradation-adaptive restoration with task feedback refinement, TaskIR balances restoration quality and downstream task performance under diverse degradation conditions.

The main contributions of this work are summarized as follows:
\begin{itemize}
    \item We propose TaskIR, a task-driven unified image restoration framework that combines degradation-adaptive restoration and task feedback refinement to balance restoration quality and downstream task performance under various degradation conditions.
    \item We develop DGTB that dynamically modulates feature transformations in a shared restoration network using degradation representations, enabling adaptive restoration across diverse degradations.
    \item We introduce TRFG and STFR to transform heterogeneous task features into targeted feedback and selectively refine restoration features, improving downstream task performance while minimizing interference with previously restored content.
    \item Extensive experiments across diverse image degradations and downstream vision tasks demonstrate that TaskIR effectively improves downstream task performance while maintaining high restoration quality.
\end{itemize}

\section{Related Work}
\textbf{All-in-One Image Restoration.}
All-in-One Image Restoration (AiOIR)~\citep{survey,tpgdiff,bioir,unirestorer} aims to handle multiple degradations, such as noise, blur, haze, and compression artifacts, within a single model, providing more flexible multi-degradation restoration than task-specific methods~\citep{low-light,hndiff}. Existing studies mainly improve the adaptability of unified models through degradation representation and dynamic restoration mechanisms. For example, AirNet~\citep{airnet} learns degradation representations via contrastive learning to guide restoration; PromptIR~\citep{promptir} employs learnable prompts to modulate the restoration process under different degradation conditions; and MoCE-IR~\citep{moceir} introduces a complexity-adaptive mixture-of-experts mechanism for dynamic assignment across degradation tasks. More recently, MIRAGE~\citep{mirage}, BioIR~\citep{bioir}, and BaryIR~\citep{baryir} further improve multi-degradation restoration performance and generalization from the perspectives of feature decomposition, bio-inspired visual modeling, and degradation-content decoupling, respectively. Despite these advances, existing AiOIR methods still mainly optimize pixel fidelity and visual quality, with limited attention to the suitability of restored images for downstream tasks.

\textbf{Task-Driven Image Restoration.}
Task-Driven Image Restoration (TDIR)~\citep{vrd-ir,sr4ir,dtiuie,adet-net} introduces downstream visual task constraints to make restored images more suitable for machine vision analysis. Existing studies~\citep{task-unirestore} have shown that image degradations can substantially affect the performance of tasks such as classification, semantic segmentation, and object detection, while optimizing only pixel-level or perceptual quality is insufficient for practical visual applications. To address this issue, SR4IR~\citep{sr4ir} and TDSR~\citep{tdsr} employ task-related losses to improve the recognition and detection performance of super-resolved images, respectively; MS-FODN~\citep{ms-fodn} jointly optimizes object detection and dehazing to improve detection under hazy conditions; ADeT-Net~\citep{adet-net} combines task feedback with textual instructions to perform dynamic dehazing for different downstream requirements; and DTIUIE~\citep{dtiuie} introduces task-driven constraints to improve the suitability of underwater image enhancement for recognition tasks. However, most existing methods are designed for a single degradation and are therefore difficult to balance restoration requirements and task performance under multiple degradations. Moreover, downstream task representations do not fully align with the requirements of image restoration, and indiscriminate refinement using task information may introduce irrelevant information and interfere with already restored content.
\begin{figure}[t]
    \centering
    \includegraphics[width=\linewidth]{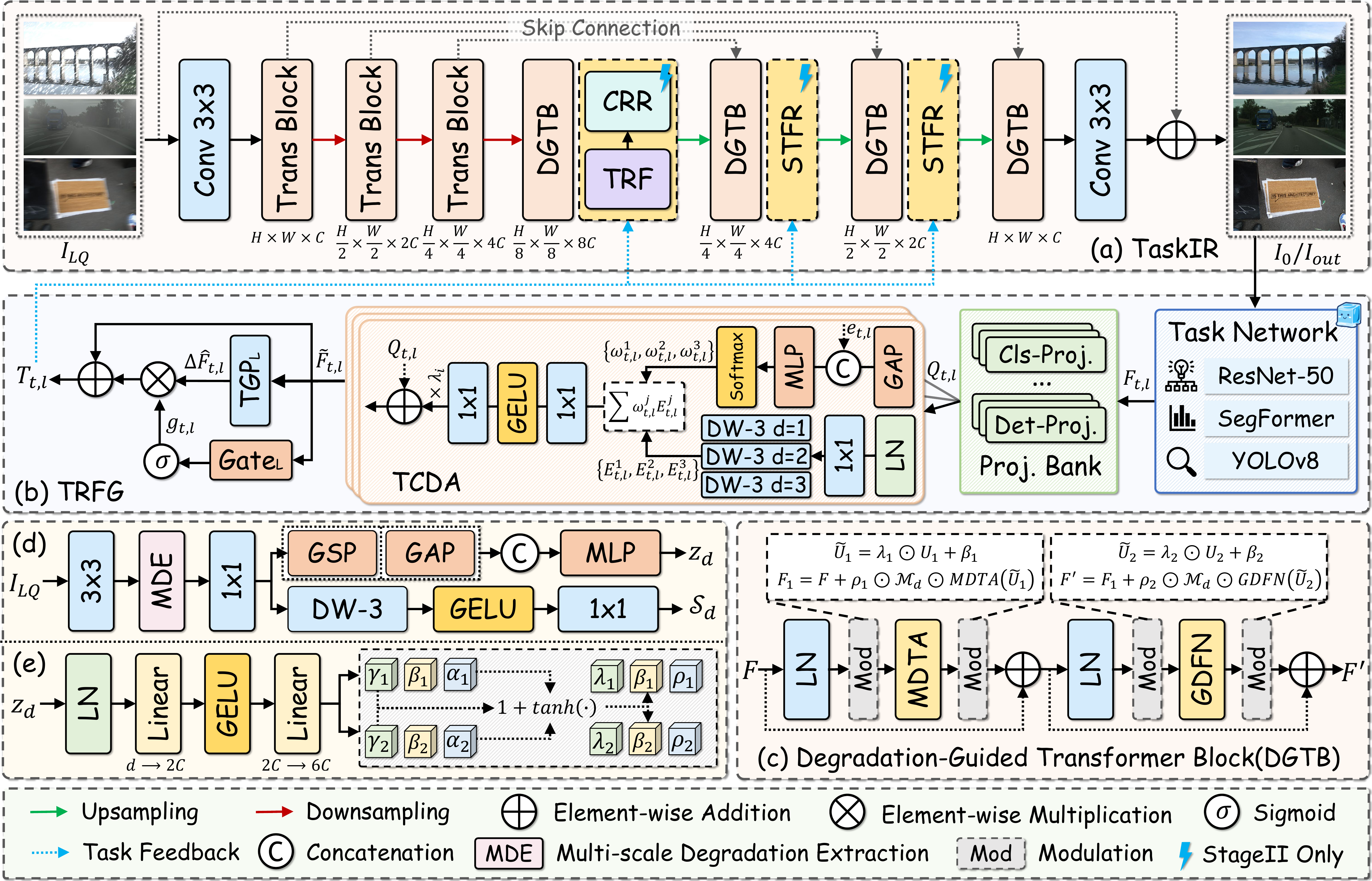}
    \caption{Overview of the proposed TaskIR framework and its key components: 
    (a) Overall architecture of TaskIR; 
    (b) Task-to-restoration feedback generation (TRFG); 
    (c) Degradation-guided transformer block (DGTB); 
    (d) Degradation representation module (DRM); and 
    (e) Degradation-conditioned parameter generator (DPG).}
    \label{fig:overall}
\end{figure}

\section{Method}
As shown in Fig.~\ref{fig:overall}, given a degraded image $I_{\mathrm{LQ}}$, TaskIR adopts a two-stage task-driven all-in-one image restoration framework to jointly improve restoration quality and downstream task performance under diverse degradations. In Stage I, the DRM extracts a global degradation representation $z_d$ and a spatial degradation response $S_d$. The DGTB then dynamically modulates the feature transformations of the shared restoration network to produce an initial restored image $I_0$. In Stage II, TRFG transforms heterogeneous downstream task features into task feedback and predicts the task-representation discrepancy of the current restored image. Based on this feedback, STFR dynamically integrates multiple task feedback signals according to their relevance to the current restoration state and selectively refines intermediate restoration features, yielding the final restored image $I_{\mathrm{out}}$.

\subsection{Stage I: Degradation-Guided Restoration}
\textbf{Degradation Representation Module.}
Different degradations exhibit distinct local spatial patterns and also induce changes in feature statistics. To characterize both types of information, we design the DRM, which generates a global degradation representation \(z_d\) and a spatial degradation response \(S_d\).
Specifically, given a degraded image \(I_{\mathrm{LQ}}\), a \(3\times3\) convolution is first applied to obtain shallow features \(X\in\mathbb{R}^{B\times C\times H\times W}\). Subsequently, Multi-scale Degradation Extraction (MDE) employs parallel depth-wise convolution branches with different receptive fields to extract multi-scale degradation patterns, followed by feature fusion and a residual connection to obtain the degradation feature \(X_d\), as illustrated in Fig.~\ref{fig:overall}(d). To characterize degradation-induced changes in global statistics, we apply GAP and Global Standard Deviation Pooling (GSP) to \(X_d\) to capture its mean and dispersion, respectively. The resulting statistics are concatenated and mapped into the global degradation representation \(z_d\in\mathbb{R}^{B\times D}\) through an MLP.

In addition to image-level degradation statistics, the degradation distribution may also vary across spatial locations. We further employ a spatial mapper \(P_s(\cdot)\) to predict the spatial degradation response from \(X_d\). 
For the \(k\)-th restoration stage, \(S_d\) is resized according to the resolution of the current features and then mapped through a Sigmoid function to obtain the corresponding spatial degradation map:
\begin{equation}
S_d = P_s(X_d),
\qquad
M_d^{(k)} = 2\sigma\left(\mathrm{Resize}_k(S_d)\right).
\end{equation}
where $\sigma(\cdot)$ denotes the Sigmoid function, and \(M_d^{(k)}\) denotes the spatial degradation map whose resolution matches the features at the \(k\)-th restoration stage.

\textbf{Degradation-Guided Transformer Block.}
After obtaining the global degradation representation $z_d$ and the spatial degradation map $M_d^{(k)}$, we design the DGTB, as illustrated in Fig.~\ref{fig:overall}(c). DGTB explicitly incorporates degradation information into the MDTA and GDFN sublayers of the Transformer, enabling the shared restoration network to adjust channel responses, residual update strengths, and feature update magnitudes at different spatial locations according to the current degradation state.
For the $k$-th restoration stage, a Degradation-conditioned Parameter Generator (DPG) first generates two sets of degradation modulation parameters from $z_d$. As shown in Fig.~\ref{fig:overall}(e), DPG employs normalization and two mapping layers to generate channel scales $\lambda_i^{(k)}$, channel biases $\beta_i^{(k)}$, and residual scales $\rho_i^{(k)}$ for the MDTA and GDFN branches, respectively, where $i\in\{1,2\}$. 

Given the input feature $F$ of DGTB, the first set of degradation parameters modulates the normalized features of the MDTA branch, followed by feature updates using the residual scale and the spatial degradation map of the current stage:
\begin{equation}
F_1
=
F
+
\rho_1^{(k)}
\odot M_d^{(k)}
\odot
\operatorname{MDTA}
\left(
\lambda_1^{(k)}
\odot \operatorname{LN}(F)
+
\beta_1^{(k)}
\right).
\end{equation}
The second set of degradation parameters takes the updated feature $F_1$ as input to modulate the GDFN branch, yielding the DGTB output:
\begin{equation}
F_2
=
F_1
+
\rho_2^{(k)}
\odot M_d^{(k)}
\odot
\operatorname{GDFN}
\left(
\lambda_2^{(k)}
\odot \operatorname{LN}(F_1)
+
\beta_2^{(k)}
\right).
\end{equation}
Here, $\lambda_i^{(k)}$ and $\beta_i^{(k)}$ adjust the channel scales and biases of the input features to MDTA or GDFN, while $\rho_i^{(k)}$ controls the update strength of the corresponding residual branch. The spatial degradation map $M_d^{(k)}$ further modulates the update magnitudes at different spatial locations according to the spatial distribution of degradations. We employ DGTBs in the latent and decoder stages, allowing degradation information to continuously guide feature restoration and ultimately producing the Stage I restoration result $I_0$.
\begin{figure}[t]
    \centering
    \includegraphics[width=\linewidth]{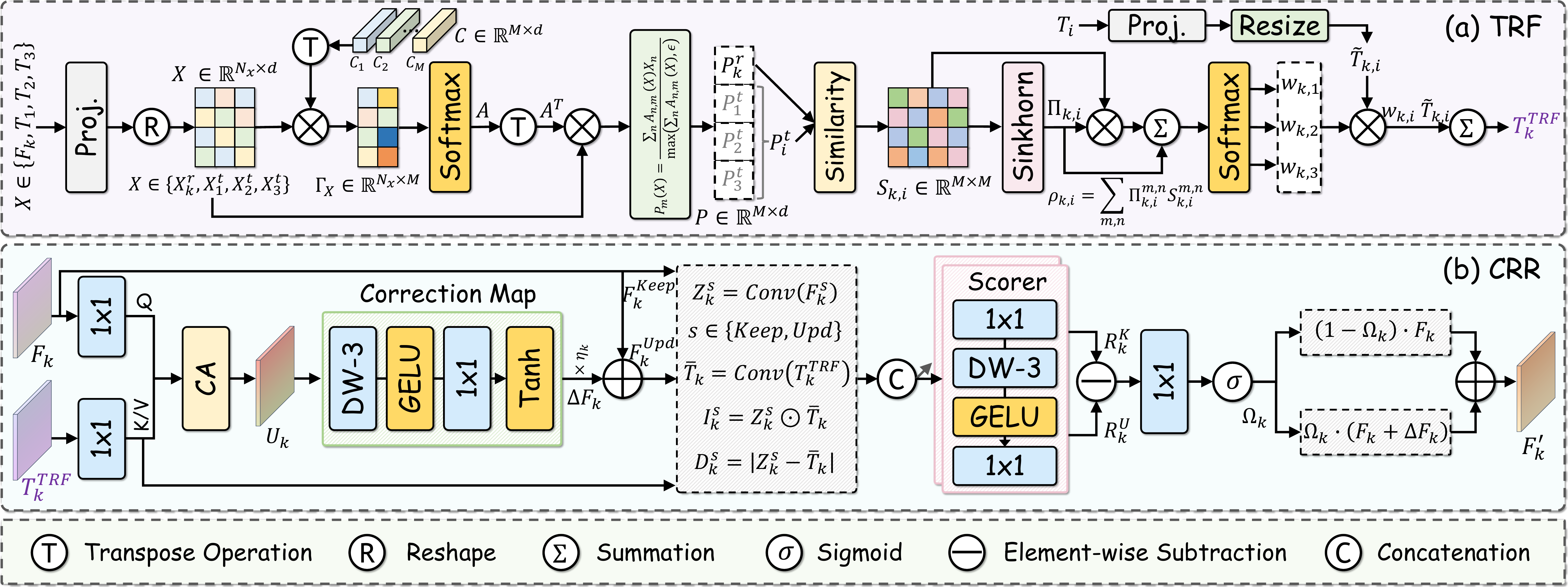}
    \caption{Architecture of the proposed selective task feedback refinement (STFR) module. 
    (a) Task-related feedback (TRF); 
    (b) Conditional restoration refinement (CRR).}
    \label{fig:stfr}
\end{figure}

\subsection{Stage II: Task-Driven Feedback Modeling}
Building upon the initial restoration results generated in Stage I, Stage II introduces downstream task feedback to further improve task performance. Specifically, we design Task-to-Restoration Feedback Generation (TRFG) and Selective Task Feedback Refinement (STFR). TRFG transforms features from different downstream tasks into task feedback, while STFR reduces interference with already restored content through feedback fusion and selective feature refinement.
\subsubsection{Task-to-Restoration Feedback Generation}
Features from different downstream tasks vary in channel dimensions, spatial scales, and representation forms. Moreover, these features are primarily designed for task prediction, making them difficult to directly apply to image restoration. To address this issue, we design TRFG to generate task feedback through adaptive feature transformation and task representation discrepancy prediction.

Given the feature $F_{t,l}$ extracted at position $l$ for task $t$, a task- and position-specific Projection Bank first performs channel mapping and spatial alignment to obtain $Q_{t,l}=P_{t,l}(F_{t,l})$, where $P_{t,l}(\cdot)$ denotes the corresponding independent projection mapping. Subsequently, the Task-Conditioned Dynamic Adapter (TCDA) employs three depth-wise convolution branches with dilation rates $r_j\in\{1,2,3\}$ to extract multi-scale information~\citep{dilated}, as illustrated in Fig.~\ref{fig:overall}(b). TCDA predicts channel-wise weights for each branch based on the global statistics of $Q_{t,l}$ and the learnable condition embedding $e_{t,l}$ associated with task $t$ and position $l$:
\begin{equation}
\omega_{t,l}^{j}
=
\operatorname{Softmax}_{j}
\left(
\operatorname{MLP}
\left(
[\operatorname{GAP}(Q_{t,l});e_{t,l}]
\right)
\right).
\end{equation}
Denoting the output of the $j$-th branch as $E_{t,l}^{j}$, the adapted task features are obtained through dynamic weighted aggregation, lightweight feature mapping, and a residual connection:
\begin{equation}
\hat{F}_{t,l}
=
Q_{t,l}
+
\xi_l
\mathcal{A}
\left(
\sum_{j=1}^{3}
\omega_{t,l}^{j}\odot E_{t,l}^{j}
\right),
\end{equation}
where $\mathcal{A}(\cdot)$ denotes the lightweight feature mapping, and $\xi_l$ is a learnable residual scale.

On this basis, the Task Gap Predictor (TGP) predicts the task representation discrepancy of the current restoration result through $\Delta\hat{F}_{t,l}=\tanh(\Phi_l(\hat{F}_{t,l}))$, where $\Phi_l(\cdot)$ denotes the discrepancy predictor corresponding to position $l$. During training, an exponential moving average (EMA) teacher~\citep{ema} is employed to construct supervision targets based on the task representation discrepancy between the initial restored image $I_0$ and the reference image $I_{\mathrm{GT}}$. During inference, the reference image is not required. Subsequently, a lightweight gate modulates the contribution of the predicted discrepancy to generate the final task feedback:
\begin{equation}
T_{t,l}
=
\hat{F}_{t,l}
+
\sigma\left(
\Psi_l(\hat{F}_{t,l})
\right)
\odot\Delta\hat{F}_{t,l},
\end{equation}
where $\Psi_l(\cdot)$ denotes the gating mapping.

\subsubsection{Selective Task Feedback Refinement}
The multiple task feedback signals generated by TRFG differ in their relevance to the current restoration state, and directly applying them to restoration features may interfere with already restored content. We therefore design STFR, as shown in Fig.~\ref{fig:stfr}, in which Task-Related Feedback (TRF) estimates the relative contributions of different feedback signals, while Conditional Restoration Refinement (CRR) further controls their actual updates to the restoration features.

\textbf{Task-Related Feedback.}
Given the restoration feature $F_k$ at the $k$-th restoration position and three task feedback signals $\{T_i\}_{i=1}^{3}$, TRF first projects them into a unified $d$-dimensional space and flattens them to obtain $X_k^r$ and $\{X_i^t\}_{i=1}^{3}$. To model the overall relationship between restoration features and task feedback, we introduce a shared prototype bank $C=\{C_m\}_{m=1}^{M}\in\mathbb{R}^{M\times d}$ containing $M$ learnable prototypes, which is shared across all restoration positions and feedback paths. For any projected feature $X$, soft assignments are computed based on its cosine similarity to the shared prototypes:
\begin{equation}
A_{n,m}(X)
=
\operatorname{Softmax}_{m}
\left(
\frac{
\langle \bar{X}_n,\bar{C}_m\rangle
}{
\tau_p
}
\right),
\end{equation}
where $\bar{X}_n$ and $\bar{C}_m$ denote the L2-normalized feature and prototype, respectively. $\operatorname{Softmax}_{m}$ denotes normalization along the prototype dimension, and $\tau_p$ denotes the prototype assignment temperature. Subsequently, the $m$-th prototype representation is obtained through weighted aggregation:
\begin{equation}
P_m(X)
=
\frac{
\sum_n A_{n,m}(X)X_n
}{
\max\left(
\sum_n A_{n,m}(X),\epsilon
\right)
}.
\end{equation}
This yields the restoration feature prototypes $P_k^r$ and the prototypes $P_i^t$ for the $i$-th task feedback signal. We then compute the cosine similarity matrix $S_{k,i}\in\mathbb{R}^{M\times M}$ between the two sets of prototypes and approximate the entropy-regularized transport plan $\Pi_{k,i}$ through log-domain Sinkhorn~\citep{sinkhorn} iterations. The relevance score between the $i$-th task feedback signal and the current restoration state, along with its contribution weight, is defined as:
\begin{equation}
\rho_{k,i}=\sum_{m=1}^{M}\sum_{n=1}^{M}\Pi_{k,i}^{m,n}S_{k,i}^{m,n},
\qquad
w_{k,i}=\operatorname{Softmax}_{i}\left(\frac{\rho_{k,i}}{\tau_r}\right).
\end{equation}
where $\operatorname{Softmax}_{i}$ denotes normalization across the three task feedback signals, and $\tau_r$ is the path weighting temperature. Each task feedback signal is then projected and resized to match the spatial resolution of the current restoration position, yielding $\widetilde{T}_{k,i}$. The aligned feedback signals are fused according to their corresponding contribution weights:
\begin{equation}
T_k^{\mathrm{TRF}}
=
\sum_{i=1}^{3}
w_{k,i}\widetilde{T}_{k,i}.
\end{equation}
Thus, TRF dynamically determines the relative contributions of different task feedback signals based on the prototype matching relationships between the current restoration state and each feedback signal, providing fused feedback for subsequent feature refinement.

\textbf{Conditional Restoration Refinement.}
Given $T_k^{\mathrm{TRF}}$, CRR further controls the extent to which task feedback modifies the current restoration features. First, channel-wise cross-attention is employed to establish interactions between restoration features and task feedback, where the query is projected from $F_k$, while the key and value are projected from $T_k^{\mathrm{TRF}}$, yielding a candidate refinement:
\begin{equation}
\Delta F_k
=
\eta_k
\tanh\left(
\mathcal{H}_k
\left(
\operatorname{CA}_k(F_k,T_k^{\mathrm{TRF}})
\right)
\right),
\end{equation}
where $\mathcal{H}_k(\cdot)$ consists of a $3\times3$ depth-wise convolution, GELU, and a $1\times1$ convolution, and $\eta_k$ is a learnable refinement scale.
Based on $\Delta F_k$, we construct a keep state $F_k^K=F_k$ and an update state $F_k^U=F_k+\Delta F_k$. The two candidate states and the fused feedback are mapped into a unified feature space, yielding $Z_k^s$ and $\bar{T}_k$, respectively. A shared Scorer then computes responses based on the conditional relationship between each candidate state and the fused feedback:
\begin{equation}
R_k^s
=
\mathcal{S}_k
\left(
[Z_k^s;\bar{T}_k;
Z_k^s\odot\bar{T}_k;
|Z_k^s-\bar{T}_k|]
\right),
\quad s\in\{K,U\}.
\end{equation}

An update gate is then generated based on the response difference between the update and keep states:
\begin{equation}
\Omega_k
=
\sigma\left(
\mathcal{G}_k(R_k^U-R_k^K)
\right),
\end{equation}
where $\mathcal{G}_k(\cdot)$ denotes a learnable gating mapping at the $k$-th restoration position, implemented using a $1\times1$ convolution. Finally, the refined restoration features are obtained by selectively fusing the keep and update states:
\begin{equation}
F_k'
=
(1-\Omega_k)\odot F_k^K
+
\Omega_k\odot F_k^U
=
F_k+\Omega_k\odot\Delta F_k.
\end{equation}
By comparing the responses of the two candidate states under the same task feedback, CRR adaptively controls the extent of feature refinement, reducing interference with already restored content.

\section{Experiments}
We evaluate TaskIR on ImageNet-1K~\citep{imagenet}, Cityscapes~\citep{cityscapes}, and PASCAL VOC2012~\citep{voc2012} across eight degradation types in terms of both image restoration quality and downstream task performance. We further conduct comprehensive comparisons and ablation studies to validate the effectiveness of the proposed framework.
\subsection{IMPLEMENTATION DETAILS}
TaskIR adopts a two-stage training strategy. In Stage I, the degradation-guided restoration network is trained using a pixel-wise \(L_1\) reconstruction loss and a degradation classification loss. In Stage II, the pretrained Stage-I weights are loaded, while the restoration backbone and downstream task networks are frozen, and only TRFG and STFR are optimized. All experiments are conducted on four NVIDIA A100 GPUs with a batch size of 8. We use the AdamW~\citep{adam} optimizer with an initial learning rate of \(1.0\times10^{-4}\). Stage I and Stage II are trained for 200K and 100K iterations, respectively.
\begin{table}[t]
\centering
\caption{Overall restoration performance across eight degradation types. The best and second-best results are shown in bold and underlined, respectively.}
\label{tab:ir}
\resizebox{\linewidth}{!}{
\begin{tabular}{l c cc cc cc}
\toprule
\multirow{2}{*}{Methods}
& \multirow{2}{*}{Venue}
& \multicolumn{2}{c}{ImageNet-1K}
& \multicolumn{2}{c}{Cityscapes}
& \multicolumn{2}{c}{VOC2012} \\
\cmidrule(lr){3-4}
\cmidrule(lr){5-6}
\cmidrule(lr){7-8}
&
& PSNR (dB) $\uparrow$
& SSIM $\uparrow$
& PSNR (dB) $\uparrow$
& SSIM $\uparrow$
& PSNR (dB) $\uparrow$
& SSIM $\uparrow$ \\
\midrule
AirNet~\citep{airnet}      & CVPR'22   & 25.20 & 0.7785 & 30.81 & 0.9005 & 25.30 & 0.7792 \\
PromptIR~\citep{promptir}    & NeurIPS'23& 27.37 & 0.8280 & 33.43 & 0.9290 & 27.32 & 0.8268 \\
NDR-Restore~\citep{ndr} & TIP'24    & 27.26 & 0.8261 & 33.67 & 0.9318 & 27.19 & 0.8243 \\
AdaIR~\citep{adair}       & ICLR'25   & 27.70 & 0.8330 & 33.70 & 0.9321 & 27.63 & 0.8314 \\
DFPIR~\citep{dfpir}       & CVPR'25   & 28.23 & 0.8420 & \underline{34.81} & 0.9334 & \underline{28.17} & 0.8408 \\
MoCE-IR~\citep{moceir}     & CVPR'25   & 24.23 & 0.7070 & 28.97 & 0.7977 & 24.20 & 0.7071 \\
VLU-Net~\citep{vlunet}     & CVPR'25   & 27.75 & 0.8359 & 34.01 & 0.9341 & 27.67 & 0.8342 \\
BaryIR~\citep{baryir}      & TPAMI'26   & \underline{28.26} & \underline{0.8451} & 34.76 & 0.9384 & 28.17 & \underline{0.8434} \\
C2SSM~\citep{c2ssm}       & CVPR'26   & 27.32 & 0.8295 & 33.60 & 0.9345 & 27.29 & 0.8286 \\
MIRAGE~\citep{mirage}      & ICLR'26   & 27.83 & 0.8407 & 34.24 & \underline{0.9397} & 27.80 & 0.8398 \\
\midrule
\textbf{TaskIR (Ours)}
& -- & \textbf{28.31} & \textbf{0.8491}
     & \textbf{35.31} & \textbf{0.9481}
     & \textbf{28.28} & \textbf{0.8484} \\
\bottomrule
\end{tabular}
}
\end{table}
\begin{table}[t]
\centering
\caption{Overall downstream task performance (\%). The best and second-best results are shown in bold and underlined, respectively.}
\label{tab:task}
\resizebox{\linewidth}{!}{
\begin{tabular}{l c cc cc cc}
\toprule
\multirow{2}{*}{Methods} 
& \multirow{2}{*}{Venue} 
& \multicolumn{2}{c}{ImageNet-1K} 
& \multicolumn{2}{c}{Cityscapes} 
& \multicolumn{2}{c}{VOC2012} \\
\cmidrule(lr){3-4}
\cmidrule(lr){5-6}
\cmidrule(lr){7-8}
& 
& Top-1 $\uparrow$ 
& Top-5 $\uparrow$ 
& mIoU $\uparrow$ 
& Dice $\uparrow$ 
& mAP$_{50}$ $\uparrow$ 
& mAP$_{50:95}$ $\uparrow$ \\
\midrule

AirNet~\citep{airnet}       & CVPR'22 & 68.31 & 87.30 & 70.95 & 82.08 & 76.90 & 58.35 \\
PromptIR~\citep{promptir}     & NeurIPS’23 & 72.26 & 89.78 & 74.12 & 84.36 & 81.92 & 63.32 \\
NDR-Restore~\citep{ndr}  & TIP'24 & 72.15 & 90.06 & 74.08 & 84.33 & 81.65 & 63.13 \\
AdaIR~\citep{adair}        & ICLR'25 & 72.10 & 90.36 & 74.12 & 84.35 & 82.05 & 63.44 \\
DFPIR~\citep{dfpir}        & CVPR'25 & 72.13 & 90.41 & 74.05 & 84.30 & \underline{82.44} & 63.14 \\
MoCE-IR~\citep{moceir}      & CVPR'25 & 65.95 & 86.50 & 66.39 & 78.39 & 72.11 & 54.01 \\
VLU-Net~\citep{vlunet}      & CVPR'25 & 72.63 & \underline{90.56} & 74.19 & 84.40 & 82.06 & 63.65 \\
BaryIR~\citep{baryir}       & TPAMI'26 & \underline{72.95} & 90.30 & 74.17 & \underline{84.40} & 82.40 & \textbf{64.15} \\
C2SSM~\citep{c2ssm}         & CVPR'26 & 71.49 & 89.86 & 73.89 & 84.19 & 80.98 & 62.46 \\
MIRAGE~\citep{mirage}       & ICLR'26 & 72.25 & 90.43 & \underline{74.63} & 84.71 & 81.95 & 63.45 \\
\midrule
\textbf{TaskIR (Ours)} 
             & --   & \textbf{73.00} & \textbf{90.65} & \textbf{74.69} & \textbf{84.77} & \textbf{82.46} & \underline{63.98} \\
\bottomrule
\end{tabular}
}
\end{table}

\subsection{Comparisons with the State-of-the-art Methods}
\begin{figure}[t]
    \centering
    \includegraphics[width=\linewidth]{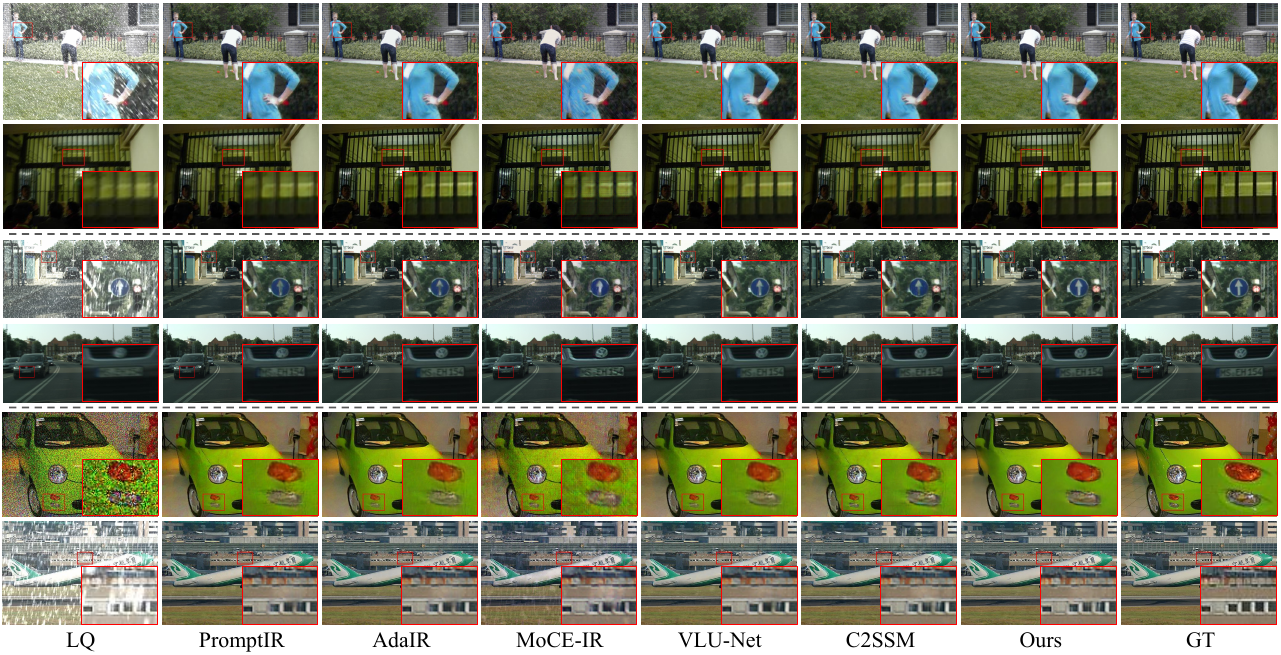}
    \caption{Qualitative comparison of image restoration results under diverse degradations on ImageNet-1K~\citep{imagenet}, Cityscapes~\citep{cityscapes}, and PASCAL VOC2012~\citep{voc2012}. Zoom in for a better view.}
    \label{fig:restoration}
\end{figure}

\textbf{Image Restoration Performance.}
Tab.~\ref{tab:ir} reports the overall image restoration performance of different methods after two-stage training. TaskIR achieves competitive PSNR and SSIM in the overall evaluation across eight degradation types, demonstrating that it maintains high restoration quality even after incorporating downstream task feedback.
Fig.~\ref{fig:restoration} further presents a qualitative comparison of image restoration results from different methods. Compared with other methods, TaskIR more effectively removes degradations and restores texture and structural details. For example, in the car headlight region, some competing methods produce blurred edges and distorted contours, whereas TaskIR better recovers the sharp boundaries and original shape of the headlight, yielding a visual result closer to the ground truth.
\begin{figure}[t]
    \centering
    \includegraphics[width=\linewidth]{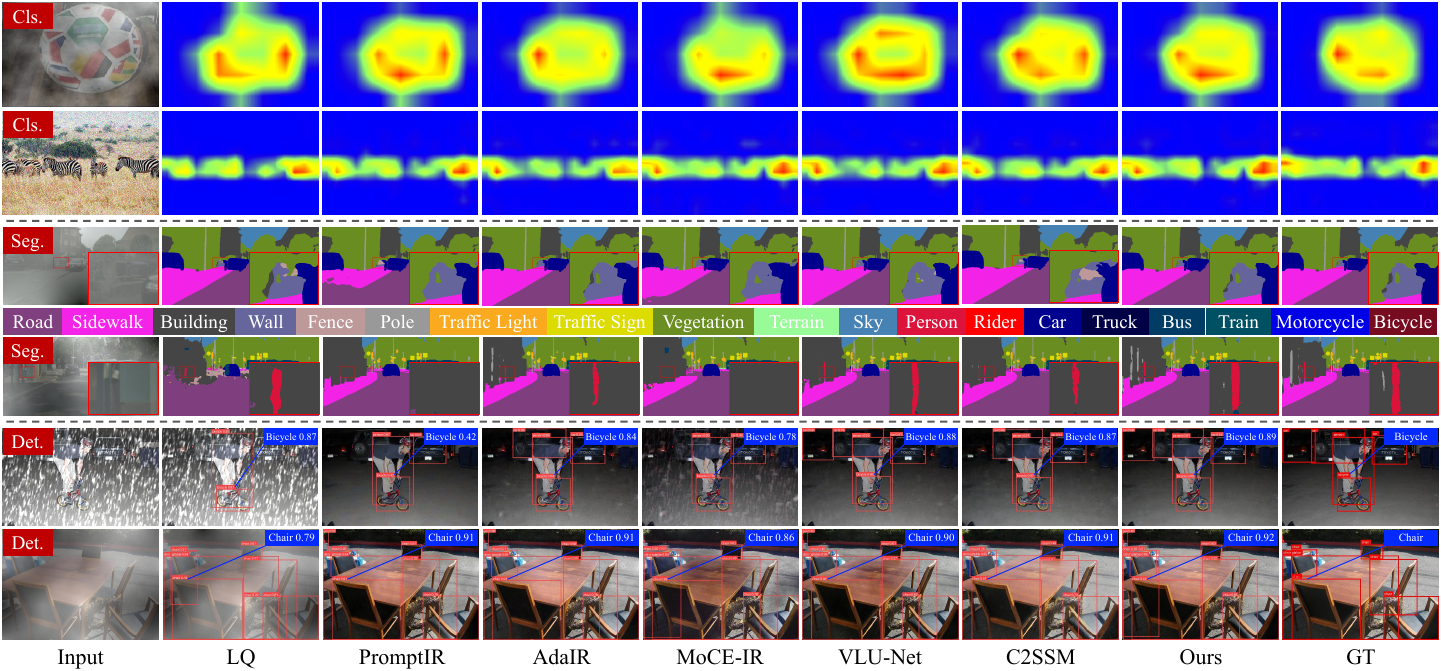}
    \caption{Qualitative comparison of downstream task results under diverse degradations, including classification (Cls.), segmentation (Seg.), and detection (Det.). Zoom in for a better view.}
    \label{fig:task}
\end{figure}

\textbf{Downstream Task Performance.}
We further evaluate the performance of the restored images on image classification, semantic segmentation, and object detection. As shown in Tab.~\ref{tab:task}, compared with other restoration methods trained with downstream task supervision, TaskIR achieves competitive performance in terms of Top-1/Top-5 accuracy~\citep{top}, mIoU/Dice~\citep{voc2012,dice}, and mAP~\citep{map}, demonstrating that its restoration results can better support diverse downstream vision tasks.
Fig.~\ref{fig:task} presents qualitative comparisons of downstream task results obtained with different methods. Compared with other methods, TaskIR produces classification predictions that are more consistent with the ground truth, segmentation results that better preserve object boundaries and local structures, and detection boxes that more closely align with target locations. For example, in building segmentation, some competing methods exhibit local misclassifications and discontinuous semantic regions, whereas TaskIR more completely preserves building regions and their boundaries, yielding better segmentation results.
\subsection{Ablation Study}
To evaluate the effectiveness of each component in TaskIR, we conduct ablation studies under the same training settings and report both image restoration quality and downstream task performance. 

\begin{figure}[t]
    \centering
    \includegraphics[width=\linewidth]{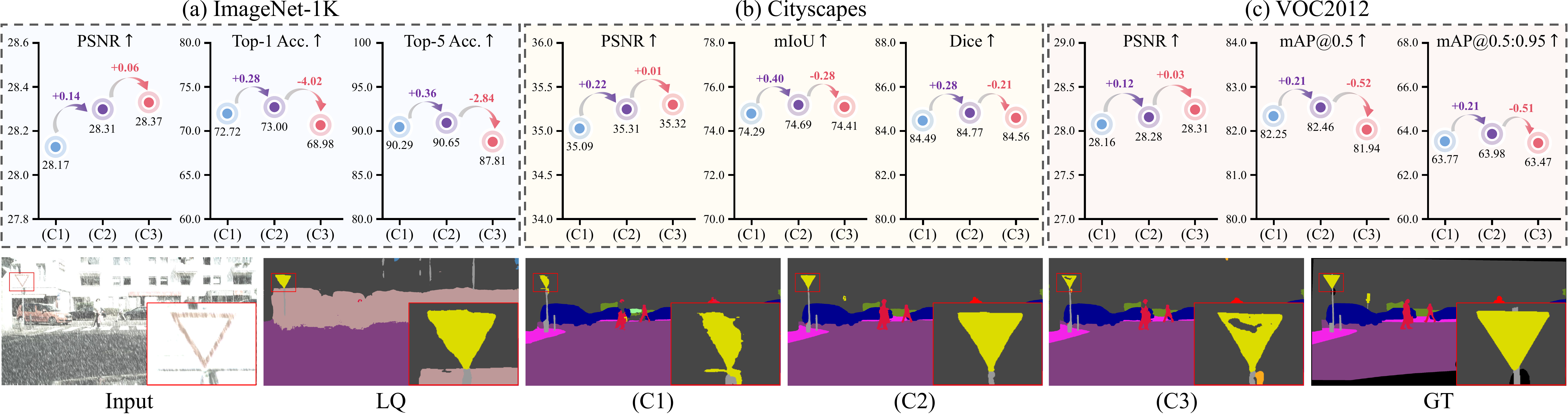}
    \caption{Ablation study on the two-stage design. Qualitative segmentation results on Cityscapes are shown for further comparison.}
    \label{fig:aba-1}
\end{figure}

\textbf{Effectiveness of the Two-Stage Design.}
We first conduct ablation studies on Stage I and Stage II to evaluate the effects of the spatial degradation map $M_d$ and task feedback, respectively. As shown in Fig.~\ref{fig:aba-1}, C1 uses only the global degradation representation $z_d$ while retaining task feedback, C2 represents the full TaskIR model, and C3 removes task feedback and directly uses the initial restored images for task prediction. Compared with C1, C2 improves restoration quality by incorporating $M_d$. Compared with C3, C2 improves downstream task performance while maintaining comparable restoration quality.

\textbf{Effectiveness of Feedback Modules.}
We further conduct ablation studies on the feedback modules in Stage II. As shown in Tab.~\ref{tab:feedback_ablation}, retaining only TRFG results in relatively low downstream task performance. Incorporating either TRF or CRR improves overall task performance, indicating that feedback fusion and selective feature refinement both contribute to more effective utilization of task feedback. Compared with configurations in which any individual module is removed, the full TaskIR model achieves better results across all downstream task metrics while maintaining comparable restoration quality.
\begin{table}[t]
\centering
\caption{Ablation study on the feedback modules. We report both restoration quality and downstream task performance across three datasets.}
\label{tab:feedback_ablation}
\resizebox{\linewidth}{!}{
\begin{tabular}{ccc cccccc}
\toprule
\multirow{2}{*}{TRFG} & \multicolumn{2}{c}{STFR} & \multicolumn{2}{c}{ImageNet-1K} & \multicolumn{2}{c}{Cityscapes} & \multicolumn{2}{c}{VOC2012} \\
\cmidrule(lr){2-3}\cmidrule(lr){4-5}\cmidrule(lr){6-7}\cmidrule(lr){8-9}
& TRF & CRR & Restoration & Classification & Restoration & Segmentation & Restoration & Detection \\
\midrule
\cmark & \xmark & \xmark & 28.29/0.8489 & 72.95/90.59 & 35.28/0.9479 & 74.52/84.64 & 28.26/0.8481 & 82.24/63.83 \\
\cmark & \cmark & \xmark & 28.28/0.8489 & 72.98/90.58 & 35.28/0.9480 & 74.65/84.74 & 28.25/0.8482 & 82.29/63.87 \\
\cmark & \xmark & \cmark & 28.30/0.8487 & 72.98/90.64 & 35.30/0.9480 & 74.67/84.75 & 28.27/0.8484 & 82.37/63.92 \\
\xmark & \cmark & \cmark & 28.31/0.8490 & 72.99/90.62 & 35.29/0.9481 & 74.64/84.73 & 28.28/0.8483 & 82.44/63.91 \\
\midrule
\cmark & \cmark & \cmark & \textbf{28.31}/\textbf{0.8491} & \textbf{73.00}/\textbf{90.65} & \textbf{35.31}/\textbf{0.9481} & \textbf{74.69}/\textbf{84.77} & \textbf{28.28}/\textbf{0.8484} & \textbf{82.46}/\textbf{63.98} \\
\bottomrule
\end{tabular}
}
\end{table}
\section{Conclusion}
In this paper, we propose TaskIR, a task-driven all-in-one image restoration framework that jointly models degradation adaptation and task feedback through a two-stage design. In Stage I, degradation representations are used to dynamically modulate Transformer feature transformations, enabling adaptive restoration across diverse degradations. In Stage II, downstream task feedback is introduced to selectively refine restoration features, enhancing their suitability for downstream tasks while reducing interference with already restored content. Extensive experiments demonstrate that TaskIR achieves competitive performance across diverse degradations and downstream vision tasks, validating the effectiveness of the proposed framework.

\subsubsection*{Acknowledgments}
This work is supported by NSFC of China under Grant 62622125, 62672410, 62301432, 62306240, 62576267, the Fundamental Research Funds for Central Universities, and Guangdong Basic and Applied Basic Research Foundation 2025A1515011119.

\bibliography{iclr2027_conference}
\bibliographystyle{iclr2027_conference}

\end{document}